\documentclass{article}
\usepackage[plainnat]{writing}

\renewcommand{\shorttitle}{Controllability and plasticity}

\title{Prior Action Control Causally Alters Language-Model Plasticity}

\author{
  Marc-Andr\'{e} Schulz\textsuperscript{1,2,3,4,\,\Letter} \\\\
  \small \textsuperscript{1}Hertie Institute for AI in Brain Health, University of T\"{u}bingen, T\"{u}bingen, Germany \\
  \small \textsuperscript{2}T\"{u}bingen AI Center, University of T\"{u}bingen, T\"{u}bingen, Germany \\
  \small \textsuperscript{3}Department of Psychiatry and Neurosciences, Charit\'{e} -- Universit\"{a}tsmedizin Berlin, Berlin, Germany \\
  \small \textsuperscript{4}German Center for Mental Health (DZPG), partner site T\"{u}bingen, T\"{u}bingen, Germany \\
}

\date{}

\usepackage{booktabs}
\usepackage{multirow}

\newcommand{\seq}[1]{\texttt{#1}}
\newcommand{\RepNStarts}{8}
\newcommand{\RepNPairs}{3}
\newcommand{\RepAllMean}{3.02}
\newcommand{\RepAllLo}{2.47}
\newcommand{\RepAllHi}{3.57}
\newcommand{\RepAllPos}{8/8}

\newcommand{\RepOppMean}{0.61}
\newcommand{\RepOppLo}{$-$0.52}
\newcommand{\RepOppHi}{1.82}

\newcommand{\RepWarning}{}

\newcommand{\DecControlSham}{3.81}
\newcommand{\DecControlShamCI}{[3.29, 4.34]}
\newcommand{\DecYokedSham}{0.56}
\newcommand{\DecYokedShamCI}{[$-$0.22, 1.50]}
\newcommand{\DecYokedShamMaxDeficit}{0.22}

\newcommand{\DecCloneSham}{3.35}

\newcommand{\SwapPRefGapMin}{$-$1.44}
\newcommand{\SwapPRefGapMax}{0.38}
\newcommand{\SwapPObsGapMin}{3.27}
\newcommand{\SwapPObsGapMax}{5.39}
\newcommand{\SwapPOwnGapMin}{3.08}
\newcommand{\SwapPOwnGapMax}{5.39}

\newcommand{\ScaleNStarts}{4}

\newcommand{\ScaleMean}{3.90}

\newcommand{\ScaleTLo}{2.29}
\newcommand{\ScaleTHi}{5.51}

\newcommand{\ScaleWarning}{}

\newcommand{\MatchedN}{16}
\newcommand{\MatchedMean}{8.67}
\newcommand{\MatchedCI}{[7.71, 9.45]}
\newcommand{\MatchedPos}{16/16}

\newcommand{\MatchedBaseline}{2.92}

\newcommand{\MatchedP}{3.05\times10^{-5}}

\begin{document}
\maketitle
\correspondence{marc-andre.schulz@uni-tuebingen.de}
\begin{abstract}
Two copies of one language model start from identical weights and receive the same rewards during training. For one copy the rewards depend on its own answers. The other copy is yoked: it receives the first copy's rewards no matter what it answers, a design taken from the learned-helplessness experiments of animal learning. We ask whether the copy whose rewards depended on its answers, the copy with control, later learns a new task faster, that is, whether control changes the model's plasticity. In Experiment 1, copies of Qwen3-0.6B generated their own answers in a two-turn game and afterwards learned a new task. The copies with control learned the new task faster in every pair, and the advantage held on different starting models as well as on the larger Qwen3-1.7B. Copies with no reward signal, trained only on the updates that every copy shared, learned about as fast as the yoked copies. The advantage is therefore mostly a benefit of control, not damage from lacking it. Control had also changed how the copies answered, so the twins practiced different answers. That practice could carry part of the advantage, and Experiment 2 removed the difference. Both copies of a pair trained on the same answers, fixed in advance, in the same order, and received the same number of rewards. Only one copy's rewards followed the answers, and that copy again learned a new task faster. The advantage of the copies with control over their yoked twins could be forecast before the new task began from each copy's weights. In sum, control over rewards during earlier training left models faster at learning a new task later on, and a conventional model evaluation need not show it.
\end{abstract}

\section{Introduction}
\label{sec:intro}

A language model's training does not end at its release.
Reinforcement learning from human feedback, fine-tuning on corrected examples, and adaptation to new domains all train a model that has already learned.
We call the speed at which a model learns from such further training its \emph{plasticity}.
A model evaluation does not measure plasticity.
An evaluation scores the answers a model gives at a single point in time, so two models that give the same answers receive the same score.
The next round of training may still change one of those models far more than the other.
We ask whether plasticity depends on one specific feature of a model's earlier training: whether the rewards the model received depended on its own answers.

To test whether rewards that depend on a model's own answers change its plasticity, we need two models whose training differs in that respect and in nothing else.
In particular, both models must have received the same rewards, so that the rewards themselves cannot explain a difference in later learning.
The \emph{yoked design} of animal learning does exactly this: it gives two animals the same outcomes while only one of them controls those outcomes.

In the classic experiment, pairs of dogs held in a harness each received a series of electric shocks \citep{seligman1967failure}.
One dog of each pair could end each shock by pressing a panel with its head.
Its partner received shocks of the same number and duration, but its panel did nothing.
The partner is called \emph{yoked}, because its outcomes are tied to another animal's actions and not to its own.
A day later each dog was placed in a box in which jumping over a low barrier ended the shock.
Dogs that had been able to end their shocks learned to escape as well as dogs that had never been shocked.
Most of the yoked dogs did not learn to escape.

In the vocabulary of animal learning, the dog that could end the shocks had \emph{control} over them: ending a shock was contingent on the dog's own action.
Seligman and Maier read the yoked dogs' failure as damage and named it \emph{learned helplessness}.
On their account the yoked dogs had learned that nothing they did mattered, and this learning undermined later attempts to escape.
\citet{maier2016helplessness} later revised this account.
Passivity under uncontrollable shock is the default response, and what the dogs with control learned was to overcome that passivity.
On the revised account the difference comes from what the dogs with control learned, not from anything the yoked dogs learned.

We port the yoked design to two copies of one language model.
Using copies removes a weakness of the yoked design that \citet{church1964yoked} identified: the two animals of a pair are different individuals.
Each shock ends when the animal with control presses, so the outcomes fit that animal's behavior better than its partner's, whatever the two individuals' differences.
Testing more pairs does not remove that bias.
Two copies of a model are identical when the experiment begins.
Only the manipulation and the random draws of answers and rewards distinguish the copies afterwards, and random draws, unlike individual differences, favor neither copy systematically.
Copies also allow two steps that animal experiments cannot take.
Both copies can be made to execute the same answers, so that only the link between answer and reward differs.
A copy then has control, in the sense of animal learning, when its rewards depend on the answers it executed, although it did not choose them.
Their weights can also be read before the new task begins.

In Experiment~1, the copy with control, which we call the \emph{contingent} copy, sampled its own answers in a two-turn game and was rewarded whenever it produced one particular sequence of two answers.
Its yoked twin received the same rewards in the same order.
After the game, each copy kept nothing but its weights, and both copies then learned a new task.
The contingent copies learned the new task faster, and the advantage held in a replication from independently trained starting models and at a larger model size.
Two copies that train on different answers end with different weights, so some difference in later learning is expected.
Two further findings show that the contingent copies' advantage is more than this expected difference.

First, the advantage could be harm to the yoked copy or benefit to the contingent copy.
Learned helplessness, the original account of the dog experiments, predicts harm to the yoked copy.
Studies of plasticity loss in neural networks point to harm as well.
Networks that train on targets whose relation to the input keeps changing lose the ability to learn new tasks \citep{lyle2023understanding,dohare2024plasticity}.
A yoked copy's rewards bear no stable relation to its answers, so its rewards push its answers in no consistent direction.
A reference arm, copies that received no reward during the game, learned the new task about as fast as yoked copies, so lacking control did little harm.
The advantage is therefore mostly a benefit of control, as the revised account of the dog experiments predicts.

Second, control changed what the contingent copies practiced: they came to produce the rewarded sequence in nearly every episode, and their twins did not.
Replaying those answers in a copy that received no rewards reproduced much of the benefit.
Experiment~2 removes that route by fixing the answers in advance.
Both copies of a pair train on the same answers, in the same order, and receive the same number of rewards.
In one copy, every reward follows one particular answer sequence, the sequence rewarded in Experiment~1.
In its twin, the rewards go to episodes chosen without regard to their answers.
The copies whose rewards followed their answers learned a new task faster.

The contingent copies' answers did not show their advantage before the new task.
In Experiment~1, on two of the four sequences the new task could ask for, they started behind their twins and still learned faster.
The advantage could nevertheless be forecast before the new task began.
A linear stand-in for each copy, built from its weights, simulated the new task's training without training the copy and predicted which twin would learn faster and by how much.

\begin{figure}[t]
  \centering
  \includegraphics[width=\linewidth]{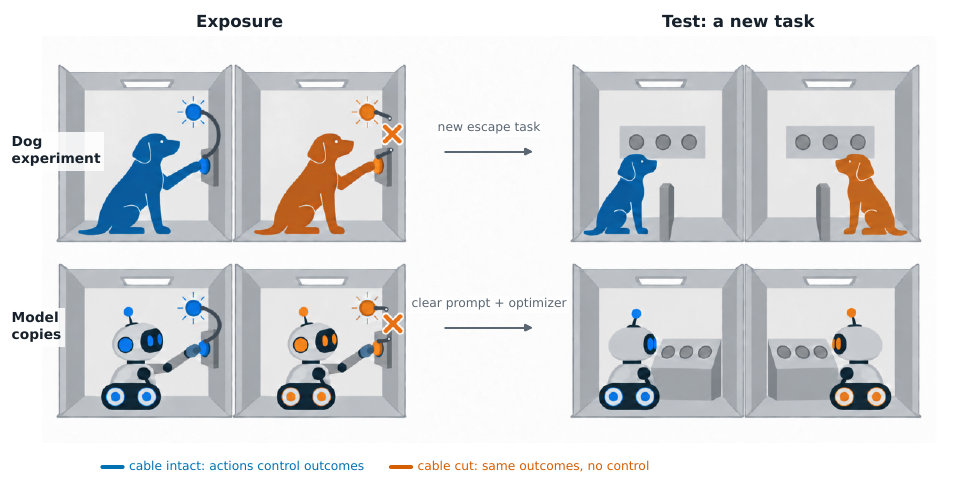}
  \caption{\textbf{The yoked design in animals and in language models.}  Exposure: two
  subjects receive the same outcomes, but only for one of them (blue) does the panel it
  presses control those outcomes.  For its yoked partner (orange) the cable is cut, and
  pressing does nothing.  Top: the learned-helplessness paradigm.  Bottom: two copies of one
  model, which keep only their weights from exposure to test.  Test: both subjects learn the
  same new task, and we compare how fast they learn it.}
  \label{fig:design}
\end{figure}

\section{Related work}
\label{sec:related}

\paragraph{Plasticity after earlier training.}
Networks that train for a long time on changing targets lose the ability to learn further tasks.
The loss is most severe when the relation between inputs and targets changes over time, and it appears even when the optimizer is reset \citep{lyle2023understanding}.
The loss also appears in continual supervised and reinforcement learning \citep{dohare2024plasticity}.
A network warm-started from training on part of a dataset generalizes worse than one trained from scratch on all of it \citep{ash2020warmstarting}.

These studies vary what a network is trained on and for how long.
None that we know of varies only how rewards are assigned to a network's outputs, with the outputs and the number of rewards matched, and then measures later learning.
In post-training, reward design decides how closely rewards follow a model's outputs, for example through how noisy a grader is.
The yoked design isolates the assignment, because yoked twins receive the same rewards and only one twin's rewards follow its answers.

\paragraph{Yoking in machine learning.}
Two studies have yoked one learning agent to another.
\citet{ostrovski2021passive} train a passive agent on the data stream that an active agent generates.
Their design follows a yoked experiment in which one kitten moved itself through an environment and its partner was carried through it \citep{held1963movement}.
The passive agent has the same architecture and loss as the active agent and sees exactly the data from which the active agent learns the task.
The passive agent still fails to learn the task.
\citet{baum2022yoking} force a candidate learning algorithm through the actions, observations, and rewards of a target agent to compare models of learning.

Both studies measure how well the yoked agent learns the task during which it is yoked.
In both, what the yoked agent lacks is actions of its own, since its data were produced by its partner's actions.
Neither study asks whether being yoked changes how an agent learns afterwards, which is a question about plasticity rather than about the yoked task.
Our copies are therefore tested on a later task.
In Experiment 2 both copies even execute the same answers, so the copy without control lacks only one thing its twin has: rewards that follow its answers.

\section{Common setup}
\label{sec:setup}

Both experiments have two phases.
In the first, \emph{exposure}, copies of one model train in pairs for 64 optimizer updates, each on a batch of eight plays of a game, under different reward conditions, which we call \emph{arms}.
The two copies of a pair are \emph{twins}, and in each experiment one twin's rewards depend on its answers while the other's do not.
In the second, the \emph{test}, we keep nothing of a copy but its weights and measure how quickly each copy learns a new task.
In both experiments, every exposure update of every copy also contains the same supervised update on a second game, which we call the rehearsal update.
Table~\ref{tab:arms} lists the comparisons and what each one tells.

\subsection{Model}
\label{sec:model}

We use Qwen3-0.6B \citep{qwen2025qwen3} and train only the MLP blocks of its last four of 28 layers, 37.7 million parameters, keeping all other weights fixed.
We chose these blocks before the experiments, because in a calibration run they learned the training tasks reliably and low-rank adapters on the attention weights did not.

\subsection{Tasks}
\label{sec:tasks}

Each task of Experiment~1 is a two-turn button game presented as text:
\begin{quote}\small
\texttt{Two-turn button game. Room: \textit{name}. Choose A or B at each turn. A reward is issued only after both turns.}\\
\texttt{Turn 1}\\
\texttt{Action:}
\end{quote}
The model answers with a single token, A or B.
The prompt then continues with ``Turn 2, Action:'', and the model answers again.
One play of the game is an \emph{episode}.
We write A as 0 and B as 1 throughout, with the first turn's answer first, so the answer sequence ``B, then A'' is written \seq{10}.
An episode has four possible answer sequences.
The answers are named by digits because some later tasks replace A and B with other tokens.
During exposure a copy samples its answers from its own probabilities.
To measure learning, we instead read the probability of each of the four possible sequences from the model's output distribution, so learning is measured without sampling error.

Each task is the game in its own room, named after \texttt{Room:} in the prompt.
The \emph{exposure task} is the game in which the arms' rewards differ.
The \emph{test task} is the game that each copy learns in the test, with new answer tokens, and no copy sees it before the test begins.
The \emph{rehearsal task} is a second game that the model learned before exposure.

Exposure does not start from the base model.
We first trained the base model with 128 supervised updates to answer \seq{10} in the exposure task and \seq{01} in the rehearsal task.
We call the result the \emph{starting checkpoint}.
We start from a model that has already learned a rule, \seq{10} in the exposure task.
Exposure then acts on a rule the model already follows, as post-training acts on behaviors a pretrained model already has.
The test targets are defined relative to that rule: the same rule, its reversal, and sequences that were never the rewarded one.
A faster gain on the same rule could come from practicing it, whereas a faster gain on the other targets cannot be relearning of the practiced sequence.

\subsection{Exposure training}
\label{sec:training}

Exposure trains each copy for 64 updates.
In each update the copy plays the exposure task eight times.
One step of the Adam optimizer \citep{kingma2015adam}, at learning rate $3\times10^{-5}$, then combines two gradients with equal weight (Supplementary Section~S1).
The first is the gradient of a reward-weighted loss over the eight episodes, which raises the probability of the sequences that earned more than the others and lowers the rest.
The second is the rehearsal update, the gradient of a supervised loss toward the answer \seq{01} in the rehearsal task, identical for every copy.
The rehearsal update is meant to keep the answer \seq{01} in the rehearsal task in place while exposure trains the exposure task.
The rehearsal update is also the one part of training that every arm shares, which makes a trained reference possible.
The reference copies of Section~\ref{sec:arms} set the reward-weighted part to zero, so they make the rehearsal update and nothing else and do no reward learning.
The experiments differ only in where the eight episodes' answers and rewards come from.
In Experiment~1 each copy samples its answers and the rewards follow the rule of its arm.
In Experiment~2 the answers are written in advance for both copies and the rewards follow a schedule.

\begin{table}[t]
\caption{The comparisons of the two experiments.}
\label{tab:arms}
\centering
\small
\begin{tabular}{p{3.4cm}p{2.6cm}p{3.6cm}p{3.6cm}}
\toprule
Comparison & Answers & Rewards & What it tells \\
\midrule
Exp.~1: contingent and yoked copy & each samples its own & the same, in the same order; only the contingent copy's follow its answers & whether control speeds later learning \\
Exp.~1: each against reference copies & each samples its own & reference copies do no reward learning & a gain from control, or a loss without it \\
Exp.~2: contingent and noncontingent copy & identical, written in advance & the same number per update; only the contingent copy's follow its answers & whether control speeds later learning with the answers fixed \\
\bottomrule
\end{tabular}
\end{table}

\subsection{Carrying only the weights forward}
\label{sec:reset}

Our question is whether exposure changes the weights in a way that affects later learning, so nothing but the weights may carry exposure into the test.
Exposure could also reach the test through the prompt or the optimizer state, and we close both channels.
We clear the prompt, so no copy can read its own history.
We reset the state of the Adam optimizer \citep{kingma2015adam}, so every copy starts the test with a fresh optimizer.
We also standardize one property of the weights, their scale.
Some rescalings of a network's weights leave its outputs unchanged but change how fast it trains \citep{dinh2017sharp,neyshabur2015pathsgd}, and exposure could change these scales differently in the two copies.
Before the test we therefore divide each hidden unit's row of the up-projection by its norm and multiply the unit's column of the down-projection by the same norm.
This rescaling leaves each copy's function unchanged, because a hidden unit's activation is linear in its up-projection (Supplementary Section~S1).

\subsection{Measuring learning speed}
\label{sec:measure}

Each copy trains on the test task toward a new target with a fresh Adam optimizer.
Its performance is measured after 0, 1, 2, 4, 8, 16, and 32 updates.
Performance is measured in terms of probability or expected reward, as defined in each experiment below.
We report differences in performance in percentage \emph{points}, so an effect of 4 points means that one copy's performance rose .04 more than its twin's.
Copies can differ before the first update.
A copy's \emph{gain} is therefore its average performance over the measurements after 1 or more updates, minus its performance before the first update.

The \emph{effect} in a pair is the gain of the copy whose rewards depended on its answers minus the gain of its twin.
A positive effect means that this copy improved more from its own start.
Each pair is tested on several targets, and its effect is the mean over those targets, so each pair contributes one number.
Confidence intervals and sign-flip tests treat each pair as one observation, because its twins share one starting checkpoint and one stream of reward draws.
When several starting checkpoints are compared (Section~\ref{sec:replication}), each checkpoint is one observation instead.
We report 95\% confidence intervals from a paired bootstrap and exact two-sided sign-flip tests over these observations.

\section{Experiment 1: each copy samples its own answers}
\label{sec:exp1}

\subsection{Design}
\label{sec:exp1design}

Experiment~1 asks whether a copy whose rewards depended on its own answers later learns faster than a twin that received the same rewards regardless of its answers.
In every exposure update, each copy samples its eight answer sequences from its own output probabilities, as in ordinary reward learning.
The contingent copy is rewarded for the sequence \seq{10}: after answering \seq{10} it receives a reward of 1 with probability .9, and after any other sequence with probability .1.
The yoked copy receives, episode by episode, the rewards its twin received, no matter what it answered.

Each copy also pays a cost of 0.2 whenever it answers B at the first turn.
The cost stands for the effort of pressing the panel in the learned-helplessness experiments \citep{seligman1967failure}.
Without it, nothing a yoked copy answered would have any consequence, and the copies could not show the dogs' pattern of pressing less.
With it, each copy controls its costs even when its rewards are its twin's.
The arms then differ in whether the rewards follow the answers, not in whether the answers have consequences at all.
The reward-weighted part of each update is a REINFORCE update \citep{williams1992reinforce}, which changes the log-probability of each sampled sequence.
The change is proportional to the episode's reward minus its cost, relative to the mean of the other seven episodes.
Eight pairs begin at the starting checkpoint and differ only in the answers their copies sampled and in the rewards drawn.

During exposure the twins came to answer differently, because the rewards followed only the contingent copy's answers.
The rewards shaped the answers: over these eight pairs, the contingent copies answered \seq{10} in 4,039 of their 4,096 episodes and the yoked copies in 634.
The yoked copies learned instead to avoid the cost, much as the yoked dogs pressed less: they produced \seq{00} in 2,369 episodes, the contingent copies in 23.
Yoking therefore holds the rewards fixed but not the answers or the costs.
A faster gain could then come from control itself, from the answers control led the contingent copies to practice, or from the yoked copies' cost learning.
The comparisons that follow separate these routes.
Reference copies net out the cost (Section~\ref{sec:arms}), and a replay of the practiced answers without their rewards measures the answers' share (Section~\ref{sec:carries}).
Experiment~2 removes both routes.

\subsection{Test}
\label{sec:exp1test}

After exposure each copy keeps only its weights (Section~\ref{sec:reset}) and is trained on the test task toward each of four target sequences in separate runs.
The four targets separate relearning of the practiced sequence from learning in general.
The \emph{same-rule} target \seq{10} is the sequence rewarded during exposure, so practice alone could speed it.
The \emph{opposite-rule} target \seq{01} reverses it, as in reversal learning \citep{izquierdo2017reversal}, so a faster gain on it can come neither from relearning nor from any tendency toward the practiced sequence.
The opposite-rule target is also the sequence every copy practiced in the rehearsal task, so every copy enters the test already producing it often (Section~\ref{sec:replication}).
The \emph{repeated-answer} targets \seq{00} and \seq{11} were never the rewarded sequence, so a faster gain on them is not relearning.

The test task uses two new answer tokens, X and Y.
Every run is repeated with the tokens swapped, so that no result depends on which token stands for which answer.
The test rewards the target sequence as exposure rewarded \seq{10}: 1 with probability .9 after the target and .1 after any other sequence.
The same cost of 0.2 applies to answering 1 at the first turn.
Each run makes 32 Adam updates at learning rate $10^{-5}$ that raise the copy's \emph{expected net reward}, its expected reward minus its expected cost, computed exactly from its answer probabilities.
Performance is this expected net reward, and we also report the \emph{target-sequence probability}, the probability of producing the target.

\subsection{Contingent copies learned the test task faster}
\label{sec:exp1result}
\label{sec:primary}

\begin{figure}[t]
  \centering
  \includegraphics[width=\linewidth]{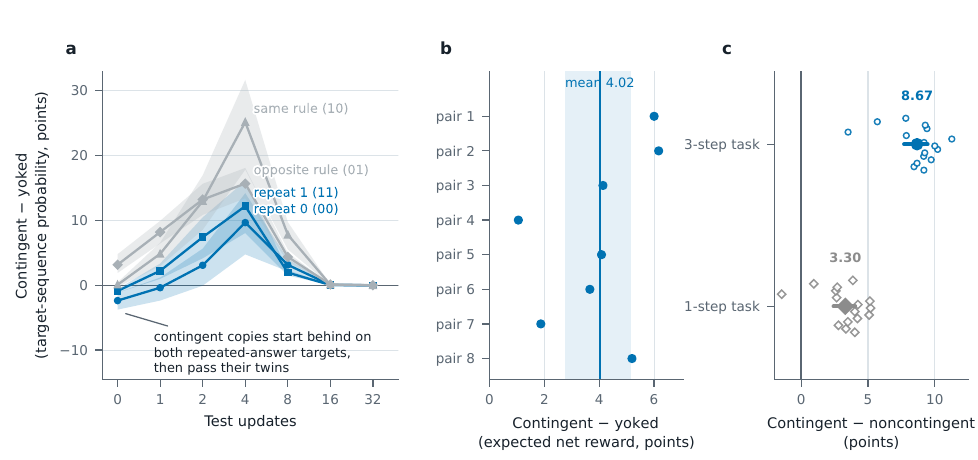}
  \caption{\textbf{Prior contingency changes later learning in both experiments.}
  \textbf{a}, Experiment 1, the first eight pairs.  The contingent copy's target-sequence
  probability minus its yoked twin's after each number of test updates, one line per
  target, averaged over the two token assignments within each pair.  Bands are paired
  bootstrap 95\% intervals over the eight pairs.  On the repeated-answer targets the
  contingent copies start behind and pass their twins by update 4.  The horizontal axis
  places the measured update counts at equal spacing, so it is not linear.
  \textbf{b}, Experiment 1: each pair's contingent-minus-yoked effect on expected net
  reward, averaged over the four targets.  The line and band are the mean and its paired
  bootstrap 95\% interval.
  \textbf{c}, Experiment 2 (Section~\ref{sec:exp2}): each pair's contingent-minus-noncontingent
  effect on the three-step test task and on a one-step task, in which a copy learns a single
  answer (Supplementary Section~S4).  Open symbols are the 16 pairs; filled symbols
  and horizontal lines are their mean and paired bootstrap 95\% interval.}
  \label{fig:results}
\end{figure}

Averaged over the four test targets, the contingent copies gained 4.02 points of expected net reward more than their yoked twins (95\% CI [2.76, 5.16]), positive in all eight pairs (exact sign-flip $p=.0078$; Supplementary Table~S2).
The gain is measured from each copy's own starting value, so a head start is not counted as learning.
On the repeated-answer targets \seq{00} and \seq{11} the contingent copies even started behind their twins, by 2.36 and 0.88 points of target-sequence probability.
After four updates they were ahead by 9.66 and 12.21 points (Figure~\ref{fig:results}a).
Both arms reached the same performance, close to the maximum, after 16 updates, so they differ in how fast they learn and not in where they end.
The effect was largest on the same-rule target, 5.39 points, where practice alone could explain it.
Control during exposure thus made the copies faster learners of the test task, also on sequences that were never the rewarded one.

\subsection{The effect replicates across starting checkpoints and model size}
\label{sec:replication}
\label{sec:scale}
\RepWarning
\ScaleWarning

All eight pairs above descend from one starting checkpoint, so the effect could be a property of that checkpoint.
We therefore trained \RepNStarts\ further starting checkpoints, each by its own random 128-update training run, and ran \RepNPairs\ contingent-yoked pairs from each, averaged within each checkpoint.
Averaged over targets, the contingent copies gained \RepAllMean\ points more than their yoked twins [\RepAllLo, \RepAllHi], positive for \RepAllPos\ checkpoints.
On the opposite-rule target alone, the sequence every copy kept practicing in the rehearsal task, the effect was small, \RepOppMean\ points [\RepOppLo, \RepOppHi] (Supplementary Section~S2).
We also ran \ScaleNStarts\ independently trained starting checkpoints of Qwen3-1.7B, a model about three times the size.
There the effect was \ScaleMean\ points, positive for all four, with an interval of [\ScaleTLo, \ScaleTHi] ($t$-interval since four checkpoints are too few for a bootstrap).
The overall effect is therefore not a property of one starting checkpoint or of one model size.

\subsection{The effect is mostly a benefit of control}
\label{sec:arms}

A positive effect has two readings: control helped the contingent copy, or the absence of control harmed the yoked one.
Telling them apart needs reference copies, which make only the rehearsal update and so do no reward learning (Section~\ref{sec:training}).
They ran on the eight further starting checkpoints, each with one new contingent and yoked pair with its own random draws.
Harm from lacking control would show as yoked copies falling below reference copies.
Instead, yoked copies gained \DecYokedSham\ points more than reference copies \DecYokedShamCI, and the interval allows them a deficit of at most \DecYokedShamMaxDeficit\ points.
The yoked copies also learned to avoid the cost, so this comparison gives the net effect of their training, uninformative rewards and an avoidable cost together.
Contingent copies gained \DecControlSham\ points more than reference copies \DecControlShamCI, on all eight checkpoints (Table~\ref{tab:reference}).
The effect is therefore mostly a benefit of control, not damage from its absence.

\begin{table}[t]
\caption{Experiment 1: the comparison with reference copies, on the eight independently trained starting checkpoints.  Each row is how much more the first arm gained than the second, in points of expected net reward, averaged over the four test targets.  The columns give the mean over the checkpoints, its bootstrap 95\% interval, and the number of checkpoints with a positive difference.  Supplementary Table~S3 gives the full comparison.}
\label{tab:reference}
\centering
\small
\begin{tabular}{lccc}
\toprule
Contrast & Mean difference & Bootstrap 95\% CI & Checkpoints with positive difference \\
\midrule
Contingent $-$ yoked & 3.24 & [2.38, 4.09] & 8/8 \\
Contingent $-$ reference & 3.81 & [3.29, 4.34] & 8/8 \\
Yoked $-$ reference & 0.56 & [$-$0.22, 1.50] & 6/8 \\
\bottomrule
\end{tabular}
\end{table}

\subsection{Replaying the practiced answers reproduced much of the benefit}
\label{sec:carries}
\label{sec:decomposition}

Control also changed what the contingent copies practiced, as the answer counts of Section~\ref{sec:exp1design} show.
A fourth arm, the \emph{matched clone}, tested how much of the benefit that practice alone reproduces.
Each clone trained by supervised updates on the answers its contingent copy had sampled, in the same order, without the episode-by-episode rewards (Supplementary Section~S3).
The clone gained \DecCloneSham\ points over reference copies, against \DecControlSham\ for the contingent copies.
In Experiment~1, much of the benefit could thus come from the answers that control led the copies to practice, the route that Experiment~2 closes.

\section{Experiment 2: both copies execute identical answers}
\label{sec:exp2}

Experiment~2 asks whether reward that follows the answers speeds later learning when the answers themselves are identical.
Both copies of a pair train on the same answer sequences, written for every episode before training, in the same order, and receive the same number of rewards in every update.
The copies differ only in which answers the rewards follow.

\subsection{Design}
\label{sec:matchedmethod}

Experiment~2 has \MatchedN\ pairs, each two copies of the starting checkpoint from which the eight pairs of Section~\ref{sec:exp1design} descend.
One copy of each pair is the \emph{contingent} copy, whose rewards follow its answers, and the other is the \emph{noncontingent} copy, whose rewards are assigned without regard to its answers.
There is no cost on any answer, because a cost that depends on the answer would give the noncontingent copy a consequence of its answers.
Each pair has its own schedule of answers and rewards, drawn at random before training.
In every update, two of the eight episodes have the sequence \seq{10}, the sequence the starting checkpoint had learned, and exactly two episodes receive a reward of 1.
Two such episodes and two rewards per update let the contingent copy's rewards follow one rule while both copies receive the same number of rewards.

The contingent copy receives its two rewards in the two episodes with the answer \seq{10}, so its rewards follow one rule: a reward after \seq{10} and after nothing else.
In the noncontingent copy the two rewards go to two episodes drawn without regard to their answers, so its rewards follow no rule.
Before training we checked that reward and answers were independent in every noncontingent schedule (Supplementary Section~S4).
Both copies train with the update rule of Experiment~1, applied to answers written in advance rather than sampled.
The update is therefore reward-weighted training on fixed answers, not an on-policy policy gradient.
Each update makes the two rewarded sequences more likely and the six others less likely.
In the contingent copy the rewarded sequence is always \seq{10}, and in the noncontingent copy it is a random one.

\subsection{Test}
\label{sec:matchedtest}

After exposure each copy again keeps only its weights and learns a test task built to share as little as possible with the exposure task.
The new task has three answer options instead of two, three answers in a row, a new prompt format and new answer tokens (Supplementary Section~S4).
No copy can therefore reuse the practiced answer sequence itself.
Unlike the test of Experiment~1, the test of Experiment~2 is supervised toward the correct sequence, with the same 32 Adam updates at learning rate $10^{-5}$.
The two tests thus cover the two kinds of further training named in the Introduction, reward-driven and supervised.
Performance is the probability of producing all three tokens of the correct sequence.

The gain is measured over a shorter window than in Experiment~1: the average performance over updates 1 to 8, minus performance before the first update.
The window covers early learning, which Experiment~2 was designed to measure.
The window was set before any exposure for Experiment~2 was generated, so it was not chosen to fit the result.
Each copy runs the test six times, with three target orders and two token sets, and a pair's effect is the mean over these runs.

\subsection{Copies whose rewards followed their answers learned the new task faster}
\label{sec:matchedresult}

The contingent copies learned the three-step task faster.
Their gain exceeded the noncontingent copies' by \MatchedMean\ points of probability \MatchedCI, positive in \MatchedPos\ pairs (exact sign-flip $p=\MatchedP$; Figure~\ref{fig:results}c).
The gain is measured from each copy's own starting value, so the contingent copies' higher start after exposure, \MatchedBaseline\ points, is not part of it.
A higher start can still change how fast probability rises, and Experiment~2 has no forecast that separates the two, as Section~\ref{sec:forecast} does for Experiment~1.
Rewards that followed the answers sped later learning even when the answers, their order and the number of rewards were identical.

\section{The advantage lies in how training moves the answers}
\label{sec:geometry}
\label{sec:forecast}
\label{sec:swap}

On two of the four targets of Experiment~1, the contingent copies started the test behind their yoked twins and still learned faster (Section~\ref{sec:exp1result}).
Ranking the copies by their answers before the test would have picked the slower learner on those targets.
Their advantage could nevertheless come from their starting answers, which shape every training step even when they are behind, or from how a training step moves those answers.
A linear stand-in for each copy separates the two, because its two parts, the starting answers and the effect of a training step, can be exchanged between copies.
For small weight changes, a network's outputs change approximately linearly with its weights, with the Jacobian as the matrix of that relation \citep{jacot2018ntk,lee2019linear,malladi2023kernelbased}.
The stand-in describes a copy by its starting answers and its Jacobian, and simulates the test's 32 updates on them without training the copy (Supplementary Section~S5).
With each copy's own Jacobian, the stand-in tracked the real test curves closely over the first eight updates, before both arms reach ceiling.
Its mean absolute error was 0.51 points of expected net reward.
It predicted effects of \SwapPOwnGapMin\ to \SwapPOwnGapMax\ points across the four targets, against observed effects of \SwapPObsGapMin\ to \SwapPObsGapMax.
These forecasts were saved before the copies were trained on the test task.
Giving both twins the Jacobian of the starting checkpoint, while each kept its own starting answers, predicted effects of \SwapPRefGapMin\ to \SwapPRefGapMax\ points, almost none.
The starting answers on their own therefore gave the contingent copies almost no advantage.
Exchanging the twins' Jacobians reversed the predicted effect, as expected if the Jacobian carries the advantage (Supplementary Section~S5).
In the stand-in, the contingent copies' advantage thus lay in how a training step moves their answers, not in the answers they started from.

\section{Discussion}
\label{sec:discussion}

How a model's rewards are assigned in the past can change how fast it learns a new task in the future.
In our experiments, two copies of one model that differed only in whether their rewards depended on their answers learned a new task at different speeds.
Animal learning calls this dependence control.
Control in this sense needs no choice by the model.
When both copies executed answers written in advance, the one whose rewards followed those answers still learned the new task faster (Section~\ref{sec:exp2}).
The dependence of rewards on answers is thus a property of training in its own right, and one that earlier work has not varied.
Research on plasticity has varied what a network trains on and for how long \citep{lyle2023understanding,dohare2024plasticity}.
Studies of post-training have found that reward learning on a model's own outputs erases less of its earlier abilities than supervised fine-tuning on fixed answers \citep{shenfeld2025razor,chen2025retaining}.
Those comparisons changed the algorithm and the data together, whereas here the data were fixed and only the dependence varied.
In post-training, the design of the rewards sets that dependence, so a choice made for the current task may also set how much training the next one needs.

The advantage was mostly a gain from control, not a loss from its absence.
In Experiment~1, yoked copies, whose rewards did not depend on their answers, learned about as fast as reference copies that did no reward learning (Section~\ref{sec:arms}).
The same number of rewards, assigned so that they followed the answers, left the contingent copies faster learners than the reference copies.
Studies of plasticity have reported losses after long training \citep{lyle2023understanding,dohare2024plasticity}, and none that we know of has varied how rewards were assigned.
Here, reward learning with control made the copies faster learners, and reward learning without it cost them little.

A benefit of control on later learning arose in a learner that has no stress physiology.
The revised account of learned helplessness also places the difference in what the subject with control learns \citep{maier2016helplessness}.
Its mechanism, though, is specific to shock: the animal with control learns to suppress its stress response.
The copies have no such mechanism and received no shocks, and the dependence of rewards on answers alone left them faster learners.
The pattern also differs: yoked dogs fell below dogs that were never shocked \citep{seligman1967failure}, while yoked copies fell at most slightly below the reference copies.
The stress mechanism accounts for the dogs' loss and cannot act in the copies.
The copies' gain is therefore a second effect of control, one that belongs to reward learning itself.
Whether animals show this second effect, a gain over subjects with no such training rather than a loss in yoked ones, is open.

How a model was rewarded in the past can change how fast it learns next, in ways its current answers do not show.
On two targets of Experiment~1 the contingent copies started behind their yoked twins and still learned faster (Section~\ref{sec:primary}).
A linear stand-in built from each copy's weights placed the advantage in how a training step moves a copy's answers, its Jacobian, rather than in the answers themselves (Section~\ref{sec:forecast}).
Choosing which model to fine-tune by its evaluation scores, or estimating from them how much training a correction will need, can therefore mislead.
Where later training matters, its speed has to be estimated directly, by a short training run or by a forecast from the weights (Section~\ref{sec:forecast}).

If the effect holds for large models, how closely a reward follows a model's outputs matters for later training, not only for what the model learns now.
In post-training, part of a reward can be unrelated to the output, for example when a grader is noisy.
A judge model that scores the same answer differently from one reading to the next adds such a part.
A verifier whose tests fail for reasons outside the answer adds one too, as does a user rating that reflects the user's mood rather than the answer.
Each makes the reward follow the output less closely, a step toward the noncontingent copies, whose rewards did not follow their answers at all.
Experiment~2 compared the two extremes, rewards that always followed the answers and rewards that never did.
We hypothesize that between them, the less a reward follows the outputs, the less that training speeds the model's later learning.
A noisy grader would then cost more than its effect on the current task.
The model would also need more training when it is next corrected or adapted, and an evaluation between the two stages need not show the difference.
The hypothesis can be tested inside a real post-training run, by reassigning the rewards at random for a growing share of the prompts.
Later learning should then slow as that share grows.

The present work's evidence comes from small models and synthetic tasks: two Qwen3 models, with only part of their weights trained, playing button games.
Building on it means yoked pairs inside real post-training runs, larger models, longer training on rewards that follow no rule, and rewarded rules the model does not already follow.
The stand-in located the advantage in how a training step moves a copy's answers, but how reward learning produces that difference remains to be explained.
A further question is whether the advantage needs rewards that follow the answers, or only training that pushes the answers consistently in one direction.
Within these limits, how a model's rewards were assigned, and not only how many it received or on which answers, changed how fast it learned what came next.

\section*{Impact statement}
A team choosing which of two trained models to fine-tune, or estimating how much training a correction will need, can compare the models' current evaluation scores.
Those scores can mislead.
How a model's training rewards were assigned changed how fast it learned from later training.
The change was encoded in its weights, where an evaluation of its answers need not reveal it.
Two models with the same evaluation scores can therefore need different amounts of training before a fine-tuning run or a correction takes hold.
The model with the better scores can even be the slower learner.
Where that matters, the response to further training has to be estimated directly, by a short trial run or by a forecast from the weights.
If the effect holds for large models, it also bears on reward design.
A grader whose rewards follow a model's outputs only in part may then cost more than the current round of training shows.
The model would also learn more slowly from the next one.

\section*{Use of AI tools}
The author designed the experiments and fixed their protocols.
An AI coding agent, Claude Code, wrote experiment and analysis code under the author's direction and turned the author's notes and revisions into draft prose, which was revised sentence by sentence.
An image model drew the illustration layer of Figure~\ref{fig:design}.

\bibliography{references}

\end{document}